\documentclass[10pt,twocolumn,letterpaper]{article}

\usepackage[pagenumbers]{cvpr} 

\usepackage[dvipsnames]{xcolor}

\usepackage{multirow}
\usepackage{color}
\usepackage{amssymb}
\usepackage{bm}
\usepackage{amsmath} 

\definecolor{cvprblue}{rgb}{0.21,0.49,0.74}
\usepackage[pagebackref,breaklinks,colorlinks,citecolor=cvprblue]{hyperref}

\def\paperID{*****} 
\def\confName{CVPR}
\def\confYear{2024}

\title{Cross-modulated Attention Transformer for RGBT Tracking}

\author{Yun Xiao$^{1}$ \;
Jiacong Zhao$^{1}$ \;
Andong Lu$^2$ \;
Chenglong Li$^{1}$ \;
Yin Lin$^3$ \;
Bing Yin$^3$ \;
Cong Liu$^3$
\\
$^1$ School of Artificial Intelligence, Anhui University, Hefei, China\\
$^2$ School of Computer Science and Technology, Anhui University, Hefei, China\\
$^3$ iFLYTEK CO.LTD., Hefei, China
}

\begin{document}
\maketitle
\begin{abstract}

Existing Transformer-based RGBT trackers achieve remarkable performance benefits by leveraging self-attention to extract uni-modal features and cross-attention to enhance multi-modal feature interaction and template-search correlation computation. 
Nevertheless, the independent search-template correlation calculations ignore the consistency between branches, which can result in ambiguous and inappropriate correlation weights. It not only limits the intra-modal feature representation, but also harms the robustness of cross-attention for multi-modal feature interaction and search-template correlation computation.
%
%
To address these issues, we propose a novel approach called Cross-modulated Attention Transformer (CAFormer), which performs intra-modality self-correlation, inter-modality feature interaction, and search-template correlation computation in a unified attention model, for RGBT tracking.
In particular, we first independently generate correlation maps for each modality and feed them into the designed Correlation Modulated Enhancement module, modulating inaccurate correlation weights by seeking the consensus between modalities.
Such kind of design unifies self-attention and cross-attention schemes, which not only alleviates inaccurate attention weight computation in self-attention but also eliminates redundant computation introduced by extra cross-attention scheme.
In addition, we propose a collaborative token elimination strategy to further improve tracking inference efficiency and accuracy.
Extensive experiments on five public RGBT tracking benchmarks show the outstanding performance of the proposed CAFormer against state-of-the-art methods.
\end{abstract}    
\section{Introduction}
\label{sec:intro}

\begin{figure}[t]
	\centering
	\includegraphics[width=0.9\linewidth]{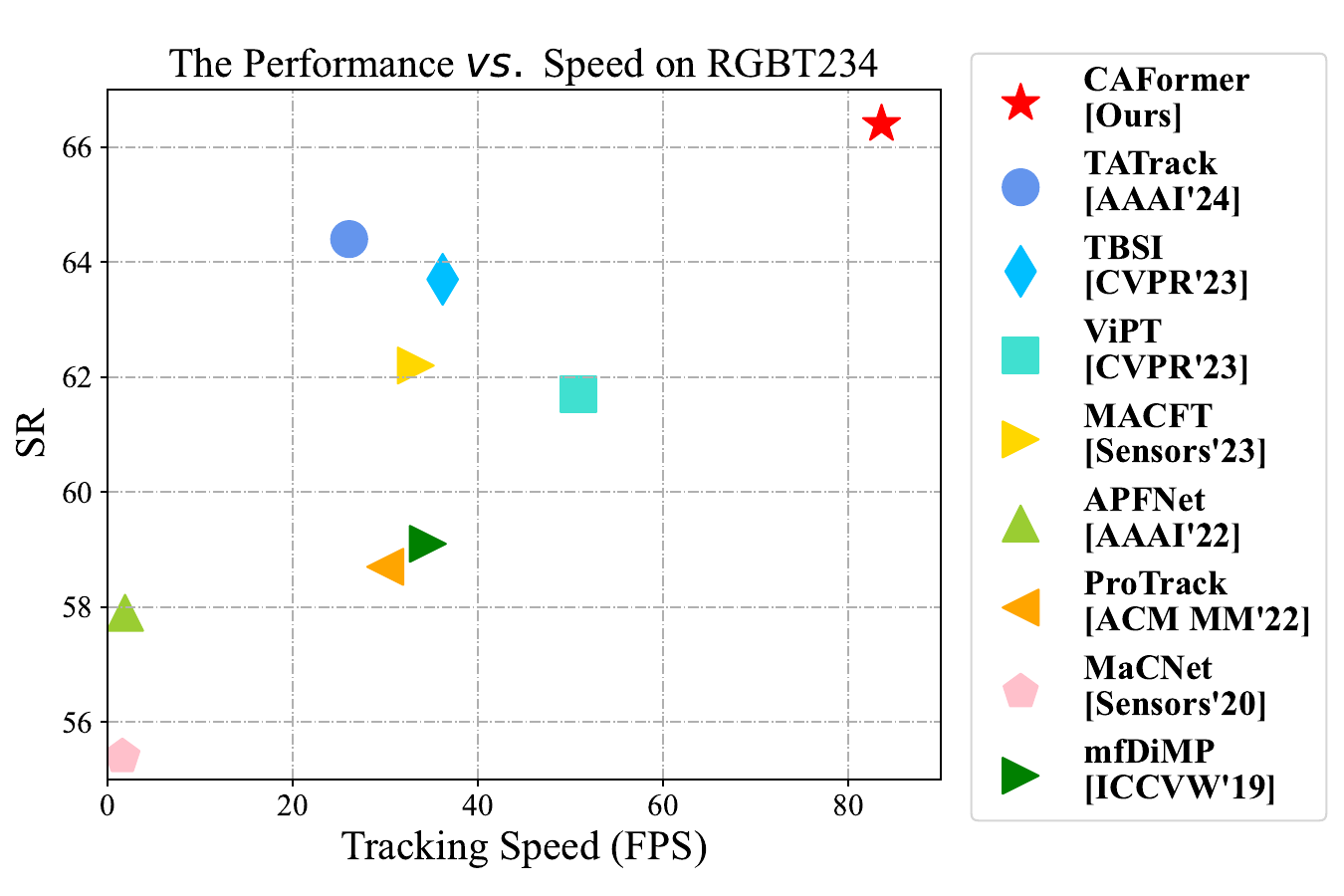}
	\caption{
        Comparison of performance and speed for state-of-the-art tracking methods on RGBT234~\cite{2019RGBT234}. We visualize the Success Rate (SR) to the Frames Per Second (FPS). Closer to the top means higher performance, and closer to the right means faster. CAFormer is able to rank the 1st in SR while running at 83.6 FPS.
	}
	\label{fig-compare}
\end{figure}

%
%

RGBT tracking~\cite{2020CAT, 2021LasHeR, 2022DMCNet, 2023TBSI, SDSTrack}, which involves fusing information from both visible and thermal infrared (TIR) modalities for visual tracking, has become an active research field in the computer vision community. 
Recently, with the success of Transformers in visual object tracking (VOT)~\cite{2021TransT, 2022ostrack, 2022simtrack}, RGBT trackers based on Transformers have gradually gained advantages in terms of speed and performance.

\begin{figure}[t]
	\centering
	\includegraphics[width=0.98\linewidth]{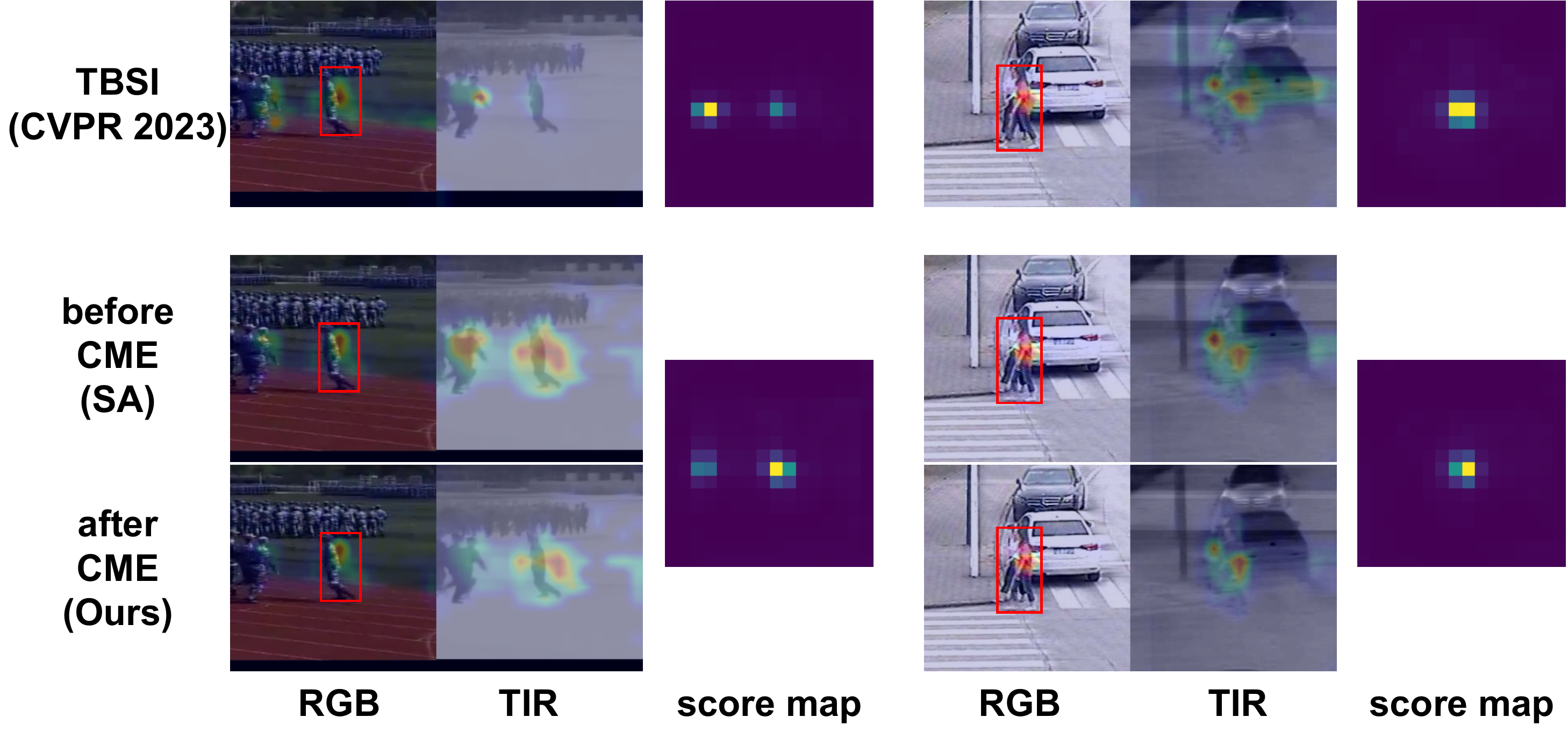}
	\caption{
        Illustration of correlation maps with different fusion methods under different modal quality inputs. The score map is the output of the location branch in the tracking head.
	}
	\label{fig:low_quality_map}
\end{figure}

The Transformer is successfully applied in RGBT tracking due to its attention mechanism, which allows it to selectively focus on relevant information and ignore irrelevant information. 
Existing Transformer-based RGBT trackers~\cite{2023TBSI, SDSTrack, 2022MIRNet} achieve remarkable performance benefits by leveraging self-attention to extract uni-modal features and cross-attention to enhance multi-modal feature interaction.
However, we observe that the calculation of correlations in self-attention is sensitive to low-quality data, resulting in ambiguous and inappropriate correlation weights, as shown in the second row of Figure~\ref{fig:low_quality_map}.
And importantly, existing works\cite{2022ostrack, 2022aiatrack, 2022SparseTT} suggest that proper correlation is important for tracking.
Therefore, we believe that there are limitations in the modality self-attention independent modeling strategy widely adopted in existing methods.
This limitation not only impairs intra-modality feature representation, but also affects subsequent multi-modal feature interactions and the robustness of template and search cross-correlation.
Moreover, existing individual computation of self-attention and cross-attention also introduces redundant computation, which limits the speed of existing RGBT trackers.

To address these issues, we propose a novel approach called Cross-modulated Attention Transformer (CAFormer), which performs intra-modality feature extraction and inter-modality feature interaction in a unified attention model, for RGBT tracking. 
Visible and infrared images in RGBT tracking are highly spatially aligned, thus their correlation between search frames and target templates is should also be consistent.
Consequently, different modality self-correlations exhibit similar interaction properties with multi-modal image features. 
%
To this end, an intuitive idea of enhancement and correction of low-quality modal correlations through high-quality modal correlations is proposed.
To adapt to the dynamic change of modal quality in RGBT tracking, a Cross-Modulated Attention (CMA) in both directions is designed to achieve adaptive correlation modulation.
%
In particular, we first compute the correlation maps for each modality independently, and then feed them into the designed Correlation Modulation Enhancement (CME) module for cross-correlation modeling to seek a correlation agreement between two modalities, which can perform the correction of inaccurate correlation relationships in previous self-attention, as shown in the third row of Figure~\ref{fig:low_quality_map}.
%
Moreover, CMA is more efficient in fusion. 
Taking ViT-Base as an example of a backbone network, for feature fusion the dimension of input features to be processed is 768.
CMA only needs to process the search-template part of the correlation map, and the dimension of the correlation vectors is related to the number of template tokens, and usually, this value is 64. By avoiding the computation of higher dimension features, CAFormer is to far outperforms existing feature fusion methods in terms of efficiency.
In summary, the proposed CMA unifies the self-attention and cross-attention schemes, which not only mitigates inaccurate correlations in self-attention, but also avoids the computational burden of additional cross-attention.
In addition, inspired by candidate eliminate method in OSTrack~\cite{2022ostrack}, we propose a collaborative token elimination strategy to further improve tracking inference efficiency and accuracy.
%
Specifically, within the search region, we consider each token as a potential candidate for the target and treat each template token as a constituent of the target object. 
Leveraging prior knowledge about the similarity between the target and each candidate provided by correlations in individual modality branches, we add the similarity of the two modalities as the overall similarity, then we kick out tokens with lower similarity.
By this way, we coordinate initial elimination results from both modalities to improve background elimination precision. 
Consequently, our module not only enhances tracking efficiency but also maintains robust performance.

Figure~\ref{fig-compare} shows the comparison of CAFormer with existing state-of-the-art methods in tracking accuracy and speed, which simultaneously achieve excellent performance in two metrics. It fully demonstrates the superiority and powerful potential of the proposed cross-modulated attention.

The contributions of this paper can be summarized as follows.
\begin{itemize}

\item We reveal the consistency exist in correlations between modalities, it brought by spatio-temporally aligned multimodal image pairs.

\item We propose a novel Cross-modulated Attention Transformer called CAFormer for accurate and efficient RGBT tracking.


\item We propose a collaborative token elimination strategy, which improves the inference efficiency with further performance enhancement. 

\item The proposed method achieves an impressive tracking speed of 83.6 FPS while achieving state-of-the-art results on three mainstream public datasets. 

\end{itemize}

\section{Related Work}

\begin{figure*}[htbp!]
	\centering
	\includegraphics[width=0.95\linewidth]{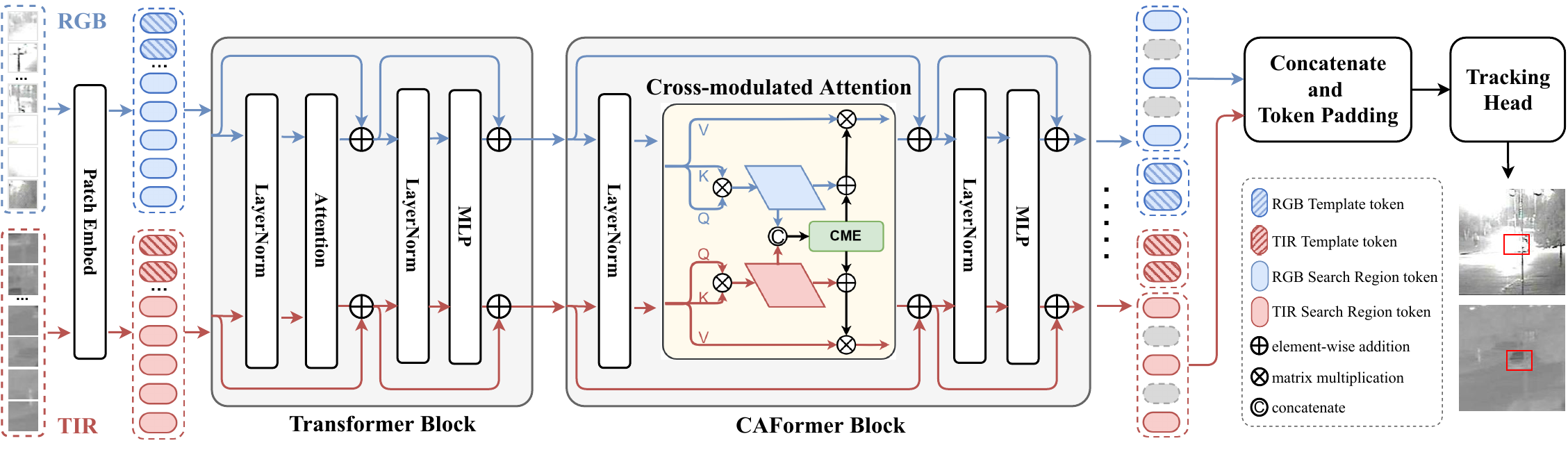}
	\caption{
		Overall framework of Cross-modulated Attention Transformer (CAFormer) for RGBT tracking. 
	}
	\label{fig:overall_framework}
\end{figure*}

\subsection{Attention Mechanism}

Attention mechanisms have been widely used in computer vision tasks over past decade \cite{hu2018squeeze, wang2018non, vaswani2017attention, guo2022attention}. Transformer \cite{vaswani2017attention} is favored among these attention mechanisms due to its powerful representation of self-attention and cross-attention. Existing attention studies can be broadly classified into two categories.
One category focuses on lightweight attention studies~\cite{liu2021swin,schlatt2024investigating, zhou2024dpnet,zhu2020deformable}. 
For example, Liu et al.~\cite{liu2021swin} reduce the computational complexity of attention by introducing local windows into self-attention. 
Schlatt et al.~\cite{schlatt2024investigating} design a sparse interaction strategy between query token and key token, which improves the efficiency of cross-attention.
However, these methods of accelerating attention may harm performance due to the remove of global relationship modelling.
Another category of studies~\cite{2022aiatrack,xu2023self} is devoted to improving the quality of attention maps. 
For example, Gao et al.~\cite{2022aiatrack} refine the original self-attention by constructing a second-order relation matrix of the self-attention map. Xu et al.~\cite{xu2023self} propose a self-calibrated cross-attention to enhance discrimination between foreground and background images.
However, these schemes are challenging to model accurately attention weights on their own information when they encounter low-quality data inputs. 
In contrast, this paper proposes a multi-modal cross-modulated attention for the first time, which enhances the attention quality of each modality by establishing a strong association between the attentions of RGB and thermal modalities.

\subsection{RGBT Tracking} 
Due to the highly complementary nature of RGB and thermal infrared (TIR) modality, using TIR modality as an additional modality can effectively improve the robustness of tracking.
Therefore, RGBT tracking is proposed and has attracted wide attention.
With the publication of large-scale RGBT datasets ~\cite{2021LasHeR, Zhang_CVPR22_VTUAV}, Transformer is widely used in RGBT tracking. 
For example, Xiao et al.~\cite{2022APFNet} design attribute-specific fusion branches and utilize Transformer to enhance attribute aggregation features and modality-specific features.
Hui et al.~\cite{2023TBSI} extend ViT~\cite{2021ViT} to a multi-modal backbone and propose using fusion templates as a medium for modal interactions to enhance feature fusion with target-related contexts.
Luo et al.~\cite{luo2023learning} employ three distinct Transformer backbones to extract both modality-specific and modality-shared features. 
Other works~\cite{OneTracker, 2023ViPT} explore the application of prompt learning to multimodal tracking.
However, the correlation calculation of each modality in these methods is performed independently, which makes it challenging to avoid inaccurate correlations for low-quality inputs, thus limiting further performance improvement.
Moreover, existing fusion modules are typically designed for high-dimensional modal features, with great demand for computational resources, which is not conducive to the goal of achieving efficient tracking.

\section{Method}

\subsection{Overview}
The proposed approach, named Cross-modulated Attention Transformer (CAFormer), is designed to address the challenges of RGBT tracking by performing intra-modality self-correlation and inter-modality feature interaction in a unified attention model. 
As illustrated in Figure~\ref{fig:overall_framework}, the framework consists of a backbone network comprising Transformer and CAFormer blocks that process flattened and embedded tokens of RGB and TIR image pairs.
The cross-modulated attention mechanism employs correlation maps from both modalities to enhance interaction in the Correlation Modulated Enhancement (CME) module.
Furthermore, to filter out non-target tokens, we employ the Collaborative Token Elimination (CTE) strategy in certain layers, which improves the reliability by add correlation maps.
Subsequently, we complete the RGB and TIR tokens belonging to the search region using a token padding scheme, and then concatenate them in the channel and feed them into the tracking head for target state prediction.

\begin{figure*}[ht]
	\centering
	\includegraphics[width=0.95\linewidth]{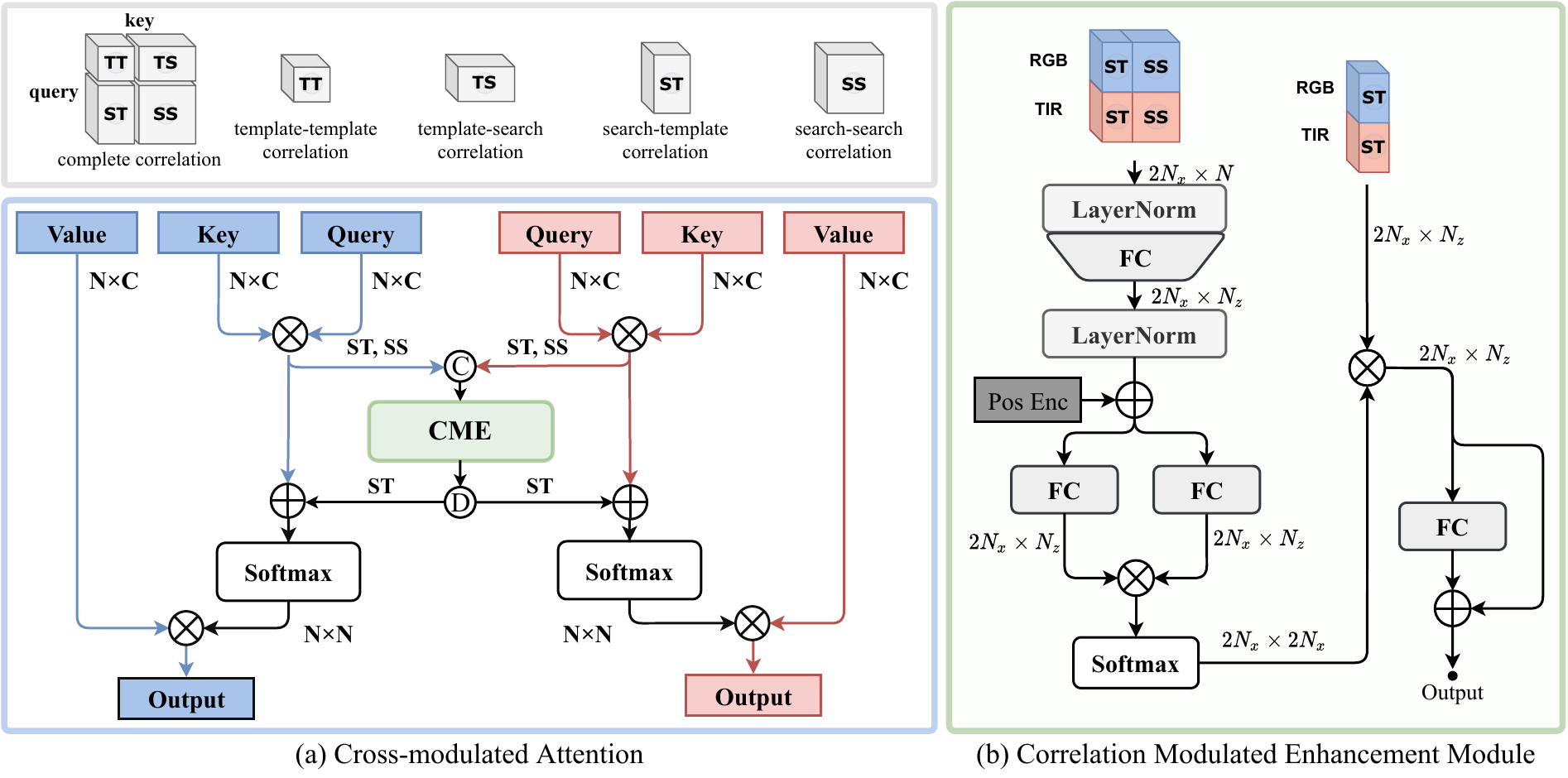}
	\caption{
    The proposed Cross-modulated Attention with the Correlation Modulated Enhancement (CME) module. 
    \includegraphics[scale=0.031]{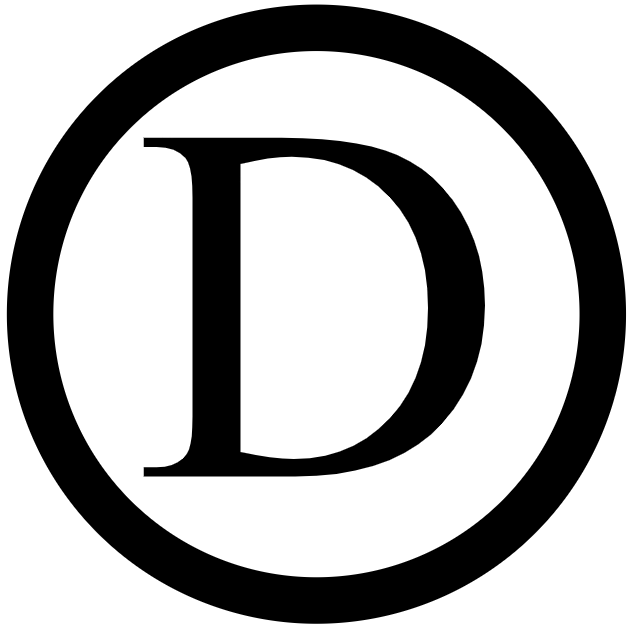} denotes dividing the features of two modalities,
    $\bigotimes$ denotes matrix multiplication, and $\bigoplus$ denotes element-wise addition. 
    The numbers beside the arrows are feature dimensions that do not include the batch size. 
    Linear projections in (a) and matrix transpose operations are omitted for brevity.
    $N$, $N_x$, and $N_z$ represent all token numbers, search region token numbers, and template region token numbers, respectively.
    }
	\label{fig-AI-location}
\end{figure*}

\subsection{RGBT Baseline Tracker} \label{chap:basetracker}
We adopt a similar approach to recent SOT (Single Object Tracking) methods~\cite{2022ostrack, 2022simtrack} by concatenating the template frames and search frames together into the Transformer backbone and then extending it to be the multi-modal backbone of our tracker.

Specifically, given the input RGB and TIR template image pair $\bm{I}_r^z, \bm{I}_t^z \in \mathbb{R}^{H_z \times W_z \times 3}$, and search region image pair $\bm{I}_r^x, \bm{I}_t^x \in \mathbb{R}^{H_x \times W_x \times 3}$ respectively,
we first divide these images into patches of size $P\times P$ and then flatten them to obtain sequences of
patches $\bm{P}_r^z, \bm{P}_t^z \in \mathbb{R}^{N_z\times (3P^2)}$ and $\bm{P}_r^x, \bm{P}_t^x \in \mathbb{R}^{N_x\times (3P^2)}$, where $N_z = H_zW_z/P^2$ and $N_x = H_xW_x/P^2$ denote the number of patches for the template and search frames, respectively.
A patch embedding layer with parameter $\bm{W}^0\in \mathbb{R}^{(3P^2)\times C}$ and learnable positional encoding $\bm{E}_z\in \mathbb{R}^{N_z \times C}$ and $\bm{E}_x\in \mathbb{R}^{N_x \times C}$ is then applied to obtain template features $\bm{Z}_r^0$, $\bm{Z}_t^0$ and search region features $\bm{X}_r^0$, $\bm{X}_t^0$ as follows:
\begin{equation}
	\begin{split}
		&\bm{Z}_r^0 = \bm{P}_r^z \bm{W}^0+\bm{E}_z,\; \bm{X}_r^0 = \bm{P}_r^x \bm{W}^0+\bm{E}_x; \\
		&\bm{Z}_t^0 = \bm{P}_t^z \bm{W}^0+\bm{E}_z,\; \bm{X}_t^0 = \bm{P}_t^x \bm{W}^0+\bm{E}_x.
	\end{split}
 \label{initial-features}
\end{equation}

Subsequently, concatenating these features yields token sequences $\bm{F}_r^0=[\bm{Z}_r^0; X_r^0]$, $\bm{F}_t^0=[\bm{Z}_t^0; X_t^0]$, then we feed them into the $l$-layer ($l=1,2,...,L$) Transformer block $T$, whose structure is shown in Figure~\ref{fig:overall_framework}. 
For simplicity, we use a tracking head consistent with OSTrack~\cite{2022ostrack} and denote it as $\phi$. 
The forward propagation process is formulated as follows:
\begin{equation}
	\begin{aligned}
		&\bm{F}_r^l=T^l(\bm{F}_r^{l-1}),\; \bm{F}_t^l=T^l(\bm{F}_t^{l-1}),\; \hfill l=1,2,...,L,\\
	\end{aligned}
  \label{propagation}
\end{equation}
where $\bm{F}_r^L, \bm{F}_t^L$ are outputs of the last Transformer block.
We merge these features along the channel dimension and feed them into the tracking head $\phi$ to derive the final predicted bounding box $\bm{B}=\phi(\bm{F}_r^L, \bm{F}_t^L)$.
At this point, we have a basic multi-modal tracker composed of two branches that share parameters and process different modalities independently.

\subsection{Cross-modulated Attention} \label{chap:CME}
Attention mechanism is a key component of the Transformer tracker~\cite{2022mixformer, 2023CTTrack}, and the correlation map is an intermediate result of the Transformer attention, which measures the similarity between the tokens~\cite{2022ostrack}.
To avoid the low-quality data affecting the correlation calculation in self-attention, we use high-quality modal correlations to achieve
enhancement and correction of low-quality modal correlations.
Considering the dynamic changes in the quality of modal correlations, a bidirectional cross-modulated strategy is used to achieve an adaptive correlation modulated process. 
We design a cross-modulated attention mechanism employs correlation maps from both modalities to enhance interaction in the Correlation Modulated Enhancement (CME) module. 

Recalling the backbone in our base tracker, the inputs to layer $l$ are $\bm{F}_r^l=[\bm{Z}_r^l; \bm{X}_r^l]$ and $\bm{F}_t^l=[\bm{Z}_t^l; \bm{X}_t^l]$, here we omit the superscript $l$ and use $\bm{F}_r$, $\bm{F}_t$ for simplicity.
$\bm{Q}_r=\bm{F}_r\bm{W}_q=[\bm{Z}_r;\bm{X}_r]\bm{W}_q=[\bm{Q}_r^z; \bm{Q}_r^x]$, $\bm{K}_r=\bm{F}_r\bm{W}_k=[\bm{Z}_r;\bm{X}_r]\bm{W}_k=[\bm{K}_r^z; \bm{K}_r^x]$ denote query and key matrix from RGB modality, and $\bm{W}_q$, $\bm{W}_k$ denote the linear projection weights for queries and keys, respectively. 
For the RGB branch, its process of RGB features to produce correlation maps $\bm{M}_r\in \mathbb{R}^{N\times N}$ can be expressed as:
\begin{equation}\label{eq:fourArea}
	\begin{aligned}
		\bm{M}_r &= \bm{Q}_r \bm{K}_r^\top = [\bm{Q}_r^z; \bm{Q}_r^x][\bm{K}_r^z; \bm{K}_r^x]^\top \\
		&= [\bm{Q}_r^z \bm{K}_r^{z\top}, \bm{Q}_r^z \bm{K}_r^{x\top};\bm{Q}_r^x \bm{K}_r^{z\top}, \bm{Q}_r^x \bm{K}_r^{x\top}]\\
		&= [\bm{M}_r^{zz}, \bm{M}_r^{zx}; \bm{M}_r^{xz}, \bm{M}_r^{xx}].
	\end{aligned}
\end{equation}
Note that $\bm{M}_r$ needs to undergo softmax and scale to be attention map in the usual meaning. 
For $\bm{Q}_t$ and $\bm{K}_t$ from TIR modality in the same way.
The processing of RGB features is symmetric to TIR features, we can get correlation maps $\bm{M}_t\in \mathbb{R}^{N\times N}$ in the same way.
As shown in Eq.~\ref{eq:fourArea}, $\bm{M}_r, \bm{M}_t$ can all be partitioned into four parts $\bm{M}_r^{zz}, \bm{M}_r^{zx}, \bm{M}_r^{xz}, \bm{M}_r^{xx}$ and $\bm{M}_t^{zz}, \bm{M}_t^{zx}, \bm{M}_t^{xz}, \bm{M}_t^{xx}$ with different roles in tracking, as proposed by~\cite{2023CTTrack}.
To simplify the description, \textbf{we named each part $\bm{TT}, \bm{TS}, \bm{ST}, \bm{SS}$ based on the query-key pairs used to calculate the correlation.}
Among them, $\bm{ST}$ is a special part, it controls the info stream from template to search frame.
Specifically, in most transformer trackers~\cite{2022ostrack, 2022simtrack}, the tracking head accepts features from the search region but actually it relies heavily on the template features to output results. Thus the effect of $\bm{ST}$ on tracking results is significant.
And importantly, due to the spatio-temporally aligned multimodal image pairs, $\bm{ST}$ within different branches have remarkable associations.

Existing methods~\cite{luo2023learning, 2023TBSI} perform separate calculations for correlation in modality, which ignores the crucial cross-modality associations.
To achieve an adaptive correlation modulated process, we design a cross-modulated attention mechanism to employ correlation maps from both modalities to enhance interaction in CME module.
The purpose of CME is to modulate $\bm{ST}$, but we need to take $\bm{SS}$ into account as well so that we can modulate the final attention map.
%
Specifically, we obtain the aggregated information $\bm{U}$ for two horizontally adjacent parts as follows:
\begin{equation}
\label{eq4}
	\begin{aligned}
		\bm{U}&=LN(LN([\bm{ST_r}, \bm{SS_r}; \bm{ST_t}, \bm{SS_t}])\bm{W}_e)\\
  &\triangleq[\bm{U}_r; \bm{U}_t]
	\end{aligned}
\end{equation}
$LN$ denotes the LayerNorm~\cite{ba2016layer} layer, and $\bm{W}_e$ is a learnable linear projection weight for embedding two correlation parts. Then we perform an attention operation on $\bm{U}$ to obtain the modulated correlation map $\bm{M}^{'}$.
\begin{equation}
	\begin{aligned}
		\bm{M}^{'} &= Softmax(\frac{(\bm{U}\bm{W}_q^{'})(\bm{U}\bm{W}_k^{'})^{\top}}{\sqrt{N_z}}) [\bm{ST_r}; \bm{ST_t}]\\
	\end{aligned}
\end{equation}
where $N_z$ is the template tokens number, 
$\bm{W}_q^{'}$ and $\bm{W}_k^{'}$ denote linear projection weights for queries and keys in CME module. Next, we separate $\bm{ST_r^{'}}$ and $\bm{ST_t^{'}}$ from the initial modulated correlation map $\bm{M}^{'}_r$ and $\bm{M}^{'}_t$, respectively.
\begin{equation}
	\begin{aligned}
		CME(\bm{M}_r; \bm{M}_t) &= \bm{M}^{'}(1+\bm{W}^{'})\\
  &= [\bm{ST_r^{'}}; \bm{ST_t^{'}}],
	\end{aligned}
\end{equation}
\begin{equation}
	\begin{aligned}
		\bm{M}_r^{'} = [0\cdot \bm{TT_r}, 0\cdot \bm{TS_r}; \bm{ST_r^{'}}, 0\cdot \bm{SS_r}],
	\end{aligned}
\end{equation}
where $\bm{W}^{'}$ is a learnable linear projection.

Finally, we add the obtained $\bm{M}_r^{'}$ to the original correlation map $\bm{M}_r$ to get the final modulated correlation map.
The process of yielding the final RGB attention map $\bm{A}_r$ can be described as follows:
\begin{equation}
	\begin{aligned}
		\bm{A}_r = Softmax(\frac{\bm{M}_r^{'}+\bm{M}_r}{\sqrt{C}}),
	\end{aligned}
\end{equation}
where $C$ denotes the dimension size of the token.

In addition, as illustrated in Figure~\ref{fig-AI-location} (a), the proposed Cross-modulated Attention is a symmetric structure in which the parameters at the corresponding positions on the left and right sides of the figure are shared.
For a multi-head attention block, we share the parameters of the CME module between the parallel multi-heads. It is worth noting that our CME module can be easily applied to other parts in the attention map.

\subsection{Collaborative Token Elimination}
Efficiency is an important metric for evaluating tracking methods~\cite{2021lighttrack, 2023mixformerv2}.
Ye et al.~\cite{2022ostrack} employ an early candidate elimination strategy to speed up the inference process in some blocks.
%
This mechanism requires constructing accurate attention weights between the target and each candidate, but it is difficult to achieve from low-quality modalities.
To solve the above problem, we propose a Collaborative Token Elimination (CTE) strategy that combines the attention weights from two modalities to make judgments.

Given the query vector $\bm{q}_r^{z}$ from $\bm{Q}_r^{z}$ and $\bm{q}_t^{z}$ from $\bm{Q}_t^{z}$ (here we follow~\cite{2022ostrack} to choose the token in the center of the template), each search region token at absolute position $i$ can be given a scalar $h_i$:
\begin{equation}
	\begin{aligned}
		\bm{h}=softmax(\bm{q}_r^{z} \bm{K}_r^{x}) + softmax(\bm{q}_t^{z} \bm{K}_t^{x}) ,
	\end{aligned}
\end{equation}
where $\bm{K}_r^{x}$ and $\bm{K}_t^{x}$ are the key vectors of search region tokens.
After that, we use $h_i$ to sort the search region tokens and keep the top-k tokens.
Our method enhances the stability of token elimination, specifically in cases where the quality of one modality declines. It accelerates the network's inference speed while maintaining robustness.

\begin{table*}[t]
	\centering
    \renewcommand{\arraystretch}{0.9} 
	\caption{Comparison with state-of-the-art methods. The top 3 results are highlighted with \textcolor{red}{red}, \textcolor{blue}{blue}, and \textcolor{green}{green} fonts, respectively. "*" indicates the speed test on the same GPU (Nvidia 3080ti).}
	\label{table_compare}
    \resizebox{\linewidth}{!}{
	\begin{tabular}{c|c|c|cc|cc|cc|ccc|cc|c}
		\hline
		\multicolumn{1}{c|}{\multirow{2}{*}{\textbf{Method}}} & \multirow{2}{*}{\textbf{backbone}} &  \multirow{2}{*}{\textbf{Pub. Info.}} & 
		\multicolumn{2}{c|}{\textbf{GTOT}} &\multicolumn{2}{c|}{\textbf{RGBT210}} &  \multicolumn{2}{c|}{\textbf{RGBT234}} & \multicolumn{3}{c|}{\textbf{LasHeR}} & \multicolumn{2}{c|}{\textbf{VTUAV}} & \multirow{2}{*}{ \textbf{FPS} $\uparrow$ } \\ 
		\multicolumn{1}{c|}{}                                  &       &                             
		&  \textbf{PR} $\uparrow$ & \textbf{SR} $\uparrow$ &  \textbf{PR} $\uparrow$ & \textbf{SR} $\uparrow$ & \textbf{PR} $\uparrow$  & \textbf{SR} $\uparrow$  & \textbf{PR} $\uparrow$  & \textbf{NPR} $\uparrow$ & \textbf{SR} $\uparrow$ &  \textbf{PR} $\uparrow$ & \textbf{SR} $\uparrow$ & \\  
        \hline	
    DAPNet~\cite{2019DAPNet} & VGG-M   & ACM MM'19 & 88.2 & 70.7 & - & - & 76.6 & 53.7 & 43.1  & 38.3  & 31.4  &  - &- &-  \\ 
	MANet~\cite{2019MANet}   & VGG-M   & ICCVW'19 & 89.4 & 72.4 & - & - & 77.7 & 53.9 & 45.5  & -     & 32.6 &- &-  & 2.1*   \\ 
	DAFNet~\cite{2019DAFNet} & VGG-M   & ICCVW'19 & 89.1 & 71.6 & - & - & 79.6 & 54.4 & 44.8 & 39.0 & 31.1 & 62.0 & 45.8 & 20.5*   \\ 
    mfDiMP~\cite{2019mfdimp} & ResNet-50 & ICCVW'19 & 83.6 & 69.7 & \textcolor{green}{84.9} & 59.3 & 84.2 & 59.1 & 59.9  & - & 46.7& 67.3 & 55.4 & \textcolor{green}{34.6*} \\	
	CAT~\cite{2020CAT}       & VGG-M   & ECCV'20 & 88.9 & 71.7 & 79.2 & 53.3 & 80.4     & 56.1 & 45.0   & 39.5  & 31.4  &- &- & - \\ 
	MaCNet~\cite{2020MaCNet} & VGG-M   & Sensors'20 & - & - & - & - & 79.0 & 55.4 & 48.2  & 42.0  & 35.0  & -&- & 1.6*   \\
	CMPP~\cite{2020CMPP} & VGG-M   & CVPR'20& \textcolor{red}{92.6} & 73.8 & - & - & 82.3 & 57.5 & - & - & - &- &- & -  \\ 
	FANet~\cite{2021FANet}   & VGG-M   & TIV'21 & - & - & - & - & 78.7 & 55.3 & 44.1  & 38.4 & 30.9 & - &-&-  \\  
	MANet++~\cite{2021MANet++}& VGG-M  & TIP'21 & 88.2 & 70.7 & - & - & 80.0 & 55.4 & 46.7  & 40.4  & 31.4  & -&- & -   \\ 
    SiamCDA~\cite{2021siamcda}& ResNet-50&TCSVT'21 & 87.7 & 73.2 & - & - & 76.0 & 56.9 & - & - & -&- &-  & - \\
	DMCNet~\cite{2022DMCNet} & VGG-M   & TNNLS'22& - & - & 79.7 & 55.5 & 83.9 & 59.3 & 49.0  & 43.1  & 35.5  &- &- & -  \\ 
	APFNet~\cite{2022APFNet} & VGG-M   & AAAI'22& 90.5 & 73.7 & - & - & 82.7 & 57.9 & 50.0  & \textcolor{green}{43.9}  & 36.2  & &- & 1.9*   \\
	MIRNet~\cite{2022MIRNet} & VGG-M & ICME'22& 90.9 & \textcolor{green}{74.4} & - & - & 81.6 & 58.9 & - & - & - &- &- & -  \\ 
	TFNet~\cite{2022tfnet} & VGG-M & TCSVT'22 & 88.6 & 72.9 & 77.7 & 52.9 & 80.6 & 56.0 & - & - & &- &- - & \\
    HMFT~\cite{Zhang_CVPR22_VTUAV}&ResNet-50& CVPR'22 & \textcolor{green}{91.2} & \textcolor{blue}{74.9} & - & - & 78.8 & 56.8 & - & - & - & \textcolor{green}{75.8} & \textcolor{green}{62.7} & 30.2 \\
    ViPT~\cite{2023ViPT}     & ViT-B & CVPR'23 & - & - & - & - & 83.5 & 61.7 & 65.1 & - & 52.5 &- &- & - \\
    MACFT~\cite{luo2023learning} & ViT-B & Sensors'23 & 90.0 & 72.7 & - & - & 85.7 & 62.2 & 65.3 & - & 51.4 & \textcolor{blue}{80.1} & \textcolor{blue}{66.8} & 33.3 \\
	TBSI~\cite{2023TBSI}     & ViT-B & CVPR'23 & - & - & \textcolor{blue}{85.3} & \textcolor{blue}{62.5} & \textcolor{green}{87.1} & 63.7  & \textcolor{green}{69.2}& \textcolor{blue}{65.7}  &\textcolor{blue}{55.6} &- &-  & \textcolor{blue}{36.2*}  \\ 
    OneTracker~\cite{OneTracker} & ViT-B & CVPR'24 & - & - & - & - & 85.7 & \textcolor{green}{64.2} & 67.2 & - & \textcolor{green}{53.8} & - & - & - \\
    Un-Track~\cite{Un-Track} & ViT-B & CVPR'24 & - & - & - & - & 84.2 & 62.5 & 66.7 & - & 53.6 &- &- & - \\
    SDSTrack~\cite{SDSTrack} & ViT-B & CVPR'24 & - & - & - & - & 84.8 & 62.5 & 66.5 & - & 53.1 &- &- & - \\
    TATrack~\cite{TATrack} & ViT-B & AAAI'24 & - & - & \textcolor{blue}{85.3} & \textcolor{green}{61.8} & \textcolor{blue}{87.2} & \textcolor{blue}{64.4} &  \textcolor{red}{70.2} & - &  \textcolor{red}{56.1} &- &- & 26.1 \\
        \hline
	CAFormer  & ViT-B & - & \textcolor{blue}{91.8} & \textcolor{red}{76.9} & \textcolor{red}{85.6} & \textcolor{red}{63.2} & \textcolor{red}{88.3} & \textcolor{red}{66.4} & \textcolor{blue}{70.0} & \textcolor{red}{66.1} & \textcolor{blue}{55.6} & \textcolor{red}{88.6} & \textcolor{red}{76.2} & \textcolor{red}{83.6*} \\
		\hline
	\end{tabular}}
\end{table*}

\section{Experiments}

\subsection{Implementation Details}
To get a more concrete understanding of the proposed method, here we present details of the implementation.
In our method, the proposed CAFormer blocks are integrated into the last 3 layers of the backbone.
The search regions are resized to 256 × 256, while the templates are resized to 128 × 128.
For the training process, CAFormer is trained on 2 NVIDIA 2080ti GPUs with a global batch size of 32. 
We set the learning rates of the backbone network and other parameters to 5e-6 and 5e-5, respectively.
The optimization algorithm employed is AdamW~\cite{loshchilov2017decoupled} with a weight decay of 1e-4. 
We train our model for 10 epochs on the training set of LasHeR~\cite{2021LasHeR}, and each epoch consists of 60K image pairs. 
For GTOT\cite{2016gtot}, RGBT210~\cite{2017RGBT210}, and RGBT234~\cite{2019RGBT234}, we directly evaluate our model without any further fine-tuning.
For the VTUAV~\cite{Zhang_CVPR22_VTUAV} dataset, we adopt the VTUAV training set for our training process, and adjust the number of training epochs to 5.
Following previous work~\cite{2023TBSI}, all experiments in this paper are loaded with pre-trained weights from the public SOT method~\cite{2022ostrack}. 

\subsection{Evaluation on Public Datasets}
Our experiments perform on five public datasets including GTOT~\cite{2016gtot}, RGBT210~\cite{2017RGBT210}, RGBT234~\cite{2019RGBT234}, LasHeR~\cite{2021LasHeR}, and VTUAV~\cite{Zhang_CVPR22_VTUAV}.
For evaluation metrics, we adopt commonly used Precision Rate (PR) and Success Rate (SR) metrics.
Following previous work~\cite{2021LasHeR}, we also adopt the Normalized Precision Rate (NPR)~\cite{muller2018trackingnet} metric for LasHeR.
In addition, GTOT~\cite{2016gtot}, RGBT234~\cite{2019RGBT234}, and VTUAV~\cite{Zhang_CVPR22_VTUAV} datasets provide the ground truth of the two modalities, following prior works~\cite{2019RGBT234, Zhang_CVPR22_VTUAV, SDSTrack} we use the best results of two modalities to circumvent small alignment errors.


\textbf{GTOT}~\cite{2016gtot} contains 50 video sequence pairs. 
As shown in Table~\ref{table_compare}, compared to previous state-of-the-art trackers~\cite{Zhang_CVPR22_VTUAV, 2020CMPP}, our method outperforms HMFT~\cite{Zhang_CVPR22_VTUAV} and the SR score is higher than CMPP~\cite{2020CMPP}.

\textbf{RGBT210}~\cite{2017RGBT210} is a challenging RGBT dataset, which contains 210 video sequence pairs, 210K frames, and 12 tracking challenge attributes.
In the evaluation of the RGBT210 dataset, our method gets the best PR/SR score with 85.6\%/63.2\%.
Compared with TBSI~\cite{2023TBSI}, there is a minor improvement of 0.3\%/0.7\% on PR/SR, but in terms of efficiency, the proposed method is twice as efficient as TBSI.
In addition, our method has a significant advantage over other methods, outperforming CAT~\cite{2020CAT}, TFNet~\cite{2022tfnet} 6.4\%/9.9\% and 7.9\%/10.3\% in terms of PR/SR, respectively.

\textbf{RGBT234}~\cite{2019RGBT234} is extended from RGBT210, which contains 12 challenge attributes, 234K frames, and 234 video sequence pairs.
As shown in Table~\ref{table_compare}, we compare our method with recently proposed RGBT trackers and achieve the best result.
TBSI~\cite{2023TBSI} is the state-of-the-art method and it uses feature fusion.
Our method outperforms TBSI by a significant margin of 1.2\%/2.7\% on PR/SR and obtains the best performance.
For other trackers, our method outperformed mfDiMP~\cite{2019mfdimp} and ViPT~\cite{2023ViPT} in PR and SR scores by 4.1\%/7.3\% and 4.8\%/4.7\%, respectively.

\textbf{LasHeR}~\cite{2021LasHeR} contains 19 challenge attributes, 734.8K frames, and 1224 video sequence pairs. 
Specifically, our tracker significantly outperforms the mfDiMP and ViPT, i.e. 10.1\%/8.9\% and 4.9\%/3.1\% respectively in PR/SR.
Although compared with TBSI~\cite{2023TBSI}, our method only has the performance advantage of 0.8\%/0.4\% in PR/NPR metrics, TBSI obviously lags behind our method in tracking efficiency because of its bulky multi-level feature interaction.

\textbf{VTUAV}~\cite{Zhang_CVPR22_VTUAV} stands out as a large-scale RGBT dataset specifically designed for UAV perspectives. 
VTUAV contains 500 video sequence pairs having 1.7M image pairs with 1920 × 1080 resolution.
It can be seen that our method outperforms all previous methods.
Specifically, compared to MACFT~\cite{luo2023learning}, which is the previous state-of-the-art method, our method leads by 8.5\%/9.4\% in PR/SR. 
This indicates that the proposed method is equally applicable to UAV scenarios and its efficiency is suitable for the needs of UAV scenarios.


\subsection{Ablation Study}

\begin{table}[tbp!]
	\centering
    \renewcommand{\arraystretch}{0.85} 
	\caption{
		Evaluation results for different structures. 
  }
	\label{table_differentStruct}
	\begin{tabular}{c|cc|cc}
		\hline    
        \multirow{2}{*}{Method} & \multicolumn{2}{c|}{RGBT234} & \multicolumn{2}{c}{LasHeR} \\
            & PR               & SR           & PR         & SR          \\ \hline
		RGBT baseline         & 86.4 &64.5 & 67.8 & 54.0   \\
		w/o $\bm{SS}$ & 87.6 &65.6  & 69.2 & 55.1 \\
		w/o Cross-modal       &  87.5&65.9  & 68.3&54.3 \\ \hline
		Full model (CAFormer) &\textbf{88.3}&\textbf{66.4}& \textbf{70.0}&\textbf{55.6} \\ \hline
	\end{tabular}
\end{table}

\subsubsection{Component Analysis}
As shown in Table~\ref{table_differentStruct}, we compare different designs for the proposed CMA module.

\textit{\textbf{w/o $\bm{SS}$.}}
Since the softmax operation will span the parts of two horizontally neighboring correlation maps, the two will affect each other.
So when we process $\bm{ST}$, we also take $\bm{SS}$ into account.
When we remove $\bm{SS}$, compared to input both $\bm{ST}$ and $\bm{SS}$ in the full model, the PR/SR score decrease by 0.7\%/0.8\% on RGBT234 and 0.8\%/0.5\% on LasHeR, respectively.
The results are shown in the Table~\ref{table_differentStruct}.
It proves that $\bm{SS}$ plays an important role in adjusting the weights of $\bm{ST}$.


\textit{\textbf{w/o Cross-modal.}}
The proposed CME module is aims to exploit the association of correlations between modalities.
When we remove this mechanism, it means that the correlation weights of the two modalities only perform self-interaction. 
As shown in the Table~\ref{table_differentStruct}, this leads to a significant decrease of 1.7\% and 1.3\% in PR and SR compared to the full model on LasHeR, respectively.
This suggests that the main performance increase of our method comes from the cross-modal interaction of correlation weights.

\subsubsection{CMA Utilization in Different Parts}
Besides interacting with $\bm{ST}$, we attempt to deploy the CME module in other parts of the correlation map.
The results on RGBT234~\cite{2019RGBT234} and LasHeR~\cite{2021LasHeR} are summarized in Table~\ref{table_different_area}, "($\bm{ST}, \bm{SS}$)" means that the two parts are considered as a whole.
We can observe the best result is obtained when applied to the $\bm{ST}$, which confirms our view that the part $\bm{ST}$ has a stronger cross-modal correlation. And the part is critical for tracking.
When we do not distinguish different parts, i.e. "($\bm{ST},\bm{SS}$)", it leads to a 1.1\%/0.9\% decrease in PR/SR scores and is less efficient compared to $\bm{ST}$.
This shows that it is necessary to distinguish different parts of the correlation map.

\begin{table}[tbp!]
	\centering
    \renewcommand{\arraystretch}{0.9} 
	\caption{Modulating different parts of the correlation map.
	}
	\label{table_different_area}
 {
	\begin{tabular}{c|cc|cc}
		\hline
        \multirow{2}{*}{part} & \multicolumn{2}{c|}{RGBT234} & \multicolumn{2}{c}{LasHeR} \\
                           & PR           & SR           & PR         & SR          \\ \hline
            $\bm{ST}$  & \textbf{88.3}&\textbf{66.4}&\textbf{70.0}& \textbf{55.6}    \\
            $\bm{SS}$  &     87.8     & 65.8         & 69.1       & 55.0    \\
            $\bm{TS}$  &     87.7     & 65.7         & 68.2       & 54.4    \\
    ($\bm{ST}$,$\bm{SS}$)& 86.4   & 64.4         & 68.9       & 54.7    \\  \hline
	\end{tabular}}
\end{table}

\begin{table}[tbp!]
	\centering
    \renewcommand{\arraystretch}{0.9} 
	\caption{
        Apply layers of the proposed CAFormer block.
	}
	\label{table_differentlayer}
	\begin{tabular}{c|cc|cc}  
		\hline
        \multirow{2}{*}{Layers} & \multicolumn{2}{c|}{RGBT234} & \multicolumn{2}{c}{LasHeR} \\
            & PR           & SR           & PR         & SR          \\ \hline
		last 1 layer     &  87.5&65.6   &   69.2&55.1  \\
		last 3 layers    &  88.3&\textbf{66.4}   &   \textbf{70.0}&\textbf{55.6}  \\ 
		last 6 layers    &  86.6&65.0   &   68.1&54.2  \\ 
		4,7,10 layers    &  \textbf{88.5}&65.8   &   69.6&55.1  \\ 
  \hline
	\end{tabular}
\end{table}

\begin{table}[tbp!]
	\centering
    \renewcommand{\arraystretch}{0.9} 
	\caption{Different candidate elimination strategies.}
	\label{table_uce}
	\begin{tabular}{cc|cc|c|c}
		\hline
		\multicolumn{2}{c|}{Method} & \multicolumn{2}{c|}{LasHeR} & \multirow{2}{*}{FPS}& \multirow{2}{*}{MACs (G)} \\
		CE  & CTE    &   PR   & SR    &    &  \\ \hline
		                        &       &  69.3 & 55.1  & 76.4 & 58.43 \\
		\checkmark &                    &  69.5 & 55.2  & 83.6 & 42.91 \\
		& \checkmark                    &  \textbf{70.0} & \textbf{55.6}  & 83.6 & 42.91 \\ \hline
	\end{tabular}
\end{table}

\subsubsection{CMA Insertion in Different Layers}
Here we insert the CAFormer block to different layers and summarize the experimental results on RGBT234~\cite{2019RGBT234} and LasHeR~\cite{2021LasHeR} in Table \ref{table_differentlayer}.
The results show that when only one CAFormer block is applied, there is already a significant improvement, which shows the necessity of correlation fusion.
When increasing to 3, the boosting effect is weakened, and when continuing to increase to 6, worse results are obtained.
This suggests that less a priori information can lead to difficulties in distinguishing potentially correct correlation weights, thus yielding erroneous interactions and resulting in performance degradation.
Finally, we choose the last 3 layers as the final solution.

\subsubsection{Different Token Elimination Schemes}
To verify the effectiveness of the proposed Collaborative Token Elimination (CTE) strategy, we also evaluate the Candidate Elimination (CE) strategy in ~\cite{2022ostrack}.
As shown in Table~\ref{table_uce}, CTE not only helps to improve the inference speed, but also significantly enhances performance, whereas the CE strategy primarily improves efficiency.
Specifically, adding the CTE or CE policy improves the tracking speed by 9.4\% and decreases the MACs by 26.6\%, while in terms of tracking performance, the CTE obtains a 0.7\%/0.5\% improvement in PR/SR, which is significantly larger than that of the CE, which is 0.2\%/0.1\%.
This indicates that the proposed method is capable of mitigating the effect of noise weights on the learning of CME modules.
And more importantly, it ensures that the weights at the corresponding locations of different modalities can interact and thus better adapt to the CME module.
We provide a comparison of the visualization results on CE and CTE in the supporting material.

\begin{figure}[tbp!]
	\centering
	\includegraphics[width=0.75\linewidth]{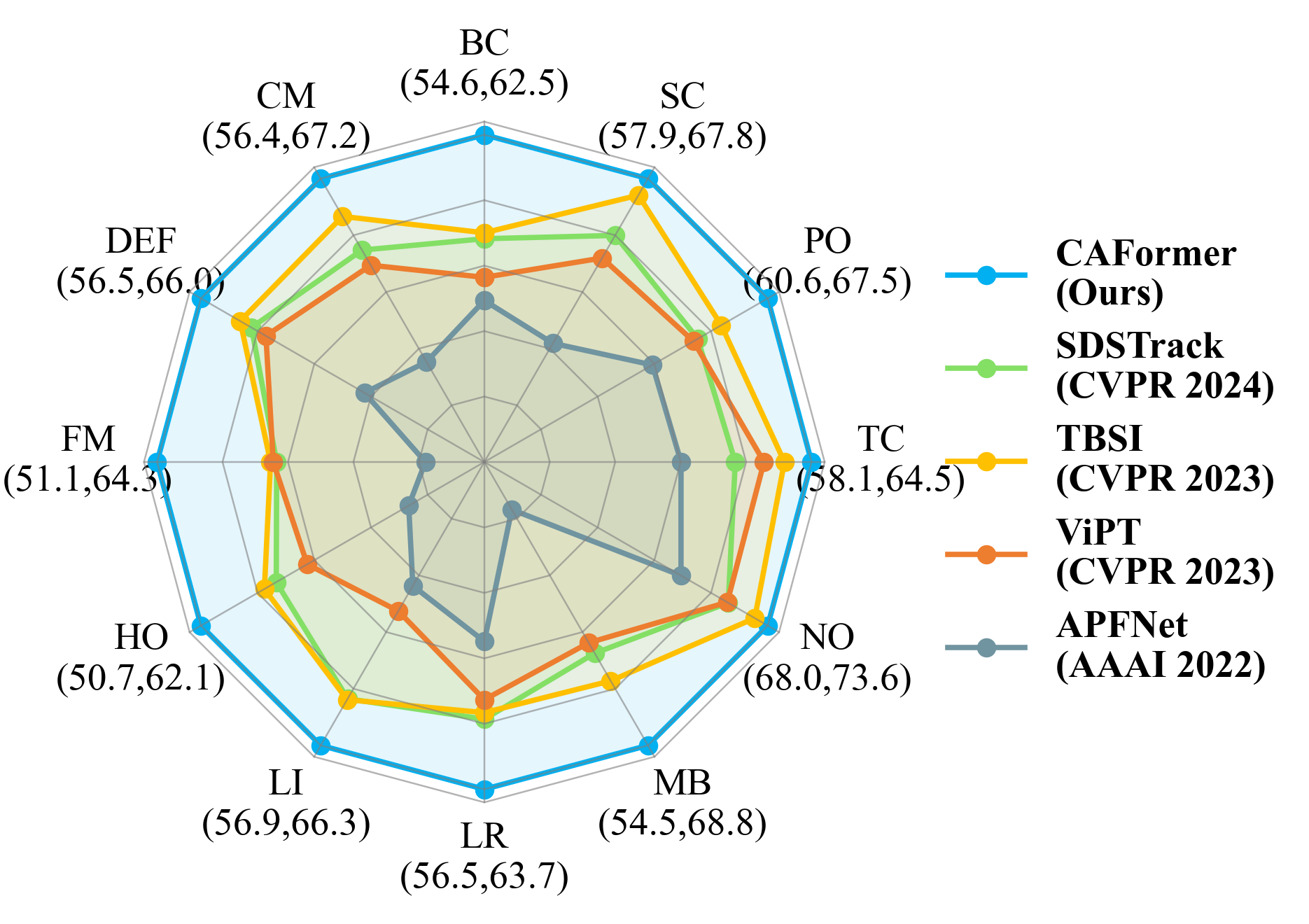}
	\caption{Attribute-based evaluation on RGBT234 dataset. In parentheses, the value on the left indicates the minimum success rate, and on the right the maximum success rate.}
	\label{fig-result}
\end{figure}

\subsection{Attribute-based Performance}
We evaluate the performance of our proposed method in various scenarios by conducting experiments on different challenge attribute subsets of the RGBT234 dataset~\cite{2019RGBT234}, including no occlusion (NO), partial occlusion (PO), heavy occlusion (HO), low illumination (LI), low resolution (LR), thermal crossover (TC), deformation (DEF), fast motion (FM), scale change (SC), motion blur (MB), camera moving (CM) and background clutter (BC).
All results are summarized in Figure~\ref{fig-result}.
Our proposed method exhibits significant improvement over the CNN-based method~\cite{2022APFNet} in challenges such as HO, MB, and others, owing to the long-range modeling capability of the Transformer.
Furthermore, our method outperforms Transformer-based methods~\cite{2023TBSI, 2023ViPT} in feature fusion under all challenges.
This demonstrates the advantages of the proposed correlation fusion scheme.

\section{Conclusion}
In this paper, we reveal a consistency in the correlations of different modal branches and exploit it to design a correlation fusion module.
The proposed method also provides a novel fusion idea for multi-modal tracking that is different from feature fusion.
Experimental results indicate that the performance of correlation fusion is competitive with or surpasses state-of-the-art feature fusion methods. 
Additionally, the paper introduces a Collaborative Token Elimination strategy that enhances the differentiation between foreground and background and further improving efficiency and performance.
In the future, we plan to combine correlation fusion and feature fusion to further improve tracking performance.

{
    \small
    \bibliographystyle{ieeenat_fullname}
    \bibliography{main}
}


\end{document}